\documentclass[10pt,twocolumn,letterpaper]{article}

\usepackage{iccv}              

\usepackage{graphicx}
\usepackage{booktabs}
\usepackage{graphicx}%
\usepackage{multirow}%
\usepackage{amsmath,amssymb,amsfonts}%
\usepackage{amsthm}%
\usepackage{mathrsfs}%
\usepackage[title]{appendix}%
\usepackage{xcolor}%
\usepackage{textcomp}%
\usepackage{manyfoot}%
\usepackage{booktabs}%
\usepackage{algorithm}%
\usepackage{algorithmicx}%
\usepackage{algpseudocode}%
\usepackage{listings}%
\usepackage{physics}
\usepackage{tikz}

\newcommand{\figref}[1]{Fig.~\ref{#1}}
\newcommand{\secref}[1]{Sec.~\ref{#1}}
\newcommand{\tabref}[1]{Table~\ref{#1}}
\makeatletter
\def\@fnsymbol#1{\ensuremath{\ifcase#1\or \dagger\or \ddagger\or
   \mathsection\or \mathparagraph\or \|\or **\or \dagger\dagger
   \or \ddagger\ddagger \else\@ctrerr\fi}}
\makeatother

\newcommand{\daggermark}{1}

\definecolor{iccvblue}{rgb}{0.21,0.49,0.74}
\usepackage[pagebackref,breaklinks,colorlinks,allcolors=iccvblue]{hyperref}

\def\paperID{2259} 
\def\confName{ICCV}
\def\confYear{2025}

\title{StyledStreets: Multi-style Street Simulator \\ with Spatial and Temporal Consistency}

\author{Yuyin Chen ~~ Yida Wang\footnotemark[\daggermark] ~~ Xueyang Zhang ~~ Kun Zhan ~~ Peng Jia ~~ Yifei Zhan ~~ Xianpeng Lang \\ \\
Li Auto Inc.\\
}

\begin{document}


\makeatletter
\g@addto@macro\@maketitle{
\begin{figure}[H]
    \setlength{\linewidth}{\textwidth}
    \setlength{\hsize}{\textwidth}
    \centering
    \includegraphics[width=\linewidth]
{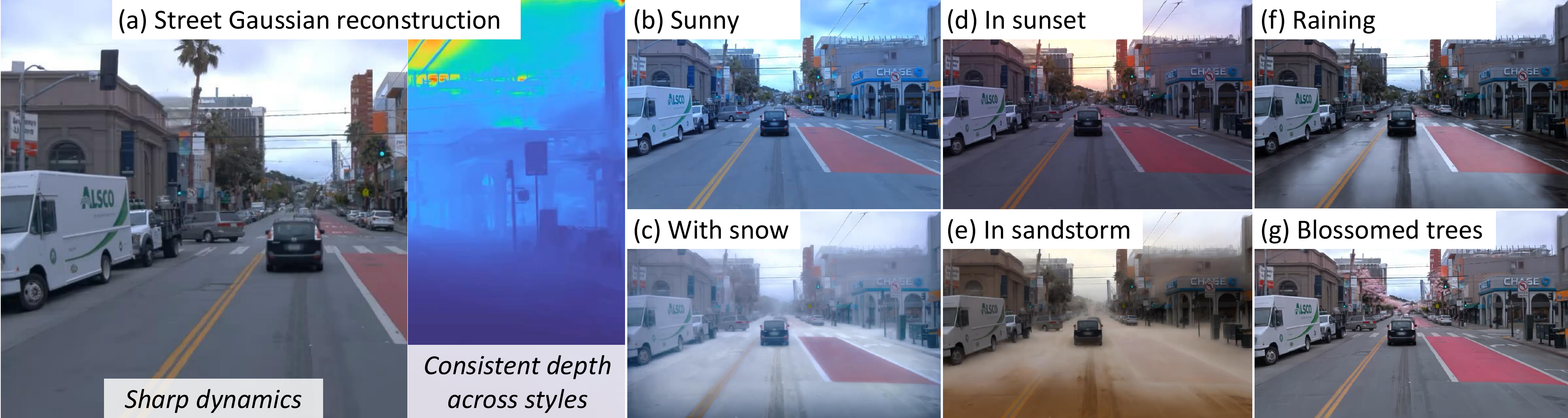}
\caption{Our proposed method migrates a pre-trained StreetGaussian model into other styles, while the geometries and dynamics are kept consistent as the original scene.}
\label{fig:styledsg}
\end{figure}
}
\maketitle

\footnotetext[1]{$\dagger$ Project lead}

\begin{abstract}
\label{sec:abstract}
Urban scene reconstruction requires modeling both static infrastructure and dynamic elements while supporting diverse environmental conditions. We present \textbf{StyledStreets}, a multi-style street simulator that achieves instruction-driven scene editing with guaranteed spatial and temporal consistency. Building on a state-of-the-art Gaussian Splatting framework for street scenarios enhanced by our proposed pose optimization and multi-view training, our method enables photorealistic style transfers across seasons, weather conditions, and camera setups through three key innovations: 
First, a hybrid embedding scheme disentangles persistent scene geometry from transient style attributes, allowing realistic environmental edits while preserving structural integrity. Second, uncertainty-aware rendering mitigates supervision noise from diffusion priors, enabling robust training across extreme style variations. Third, a unified parametric model prevents geometric drift through regularized updates, maintaining multi-view consistency across seven vehicle-mounted cameras.

Our framework preserves the original scene's motion patterns and geometric relationships. Qualitative results demonstrate plausible transitions between diverse conditions (snow, sandstorm, night), while quantitative evaluations show state-of-the-art geometric accuracy under style transfers. The approach establishes new capabilities for urban simulation, with applications in autonomous vehicle testing and augmented reality systems requiring reliable environmental consistency. Codes will be publicly available upon publication.
\end{abstract}

\section{Introduction}
\label{sec:introduction}

Urban scene reconstruction and simulation play pivotal roles in autonomous vehicle development, augmented reality applications, and digital twin creation \cite{feng2021urban}. While recent advances in neural rendering, particularly Neural Radiance Fields (NeRFs) \cite{mildenhall2021nerf}, have demonstrated impressive view synthesis capabilities, modeling dynamic urban environments with diverse stylistic variations remains challenging. These challenges stem from three fundamental requirements: 1) Simultaneous representation of persistent infrastructure and transient objects, 2) Consistent appearance editing across environmental conditions (e.g., weather, seasons), and 3) Faithful geometric reconstruction under varying camera configurations.

Current approaches face notable limitations. Static scene representations \cite{mip_splatting, nerf_w, 3d_gaussian_splatting} cannot handle moving objects, while dynamic NeRF variants struggle with computational efficiency at street scales. Recent Gaussian Splatting (3DGS) methods \cite{kerbl3Dgaussians} offer real-time rendering but lack mechanisms for style-consistent scene editing \cite{liu2024dynamicgaussians}. Multi-view systems \cite{omnire} improve geometric accuracy but cannot maintain temporal consistency during appearance modifications. This creates a critical gap in urban simulation pipelines that require both photorealistic editing and geometric reliability for downstream tasks like sensor simulation \cite{waymo2020}.

We present \textbf{\textit{StyledStreets}}, a multi-style street simulator that achieves instruction-driven scene editing with guaranteed spatial and temporal consistency. Building on 3D Gaussian Splatting \cite{kerbl3Dgaussians}, our method addresses three core challenges in urban environment modeling:

\begin{itemize}

    \item \textbf{Multi-Camera Consistency}: Vehicle-mounted camera arrays \cite{waymo2020} introduce complex cross-view constraints. We preserve inter-camera geometry through joint pose optimization (Sec.~\ref{sec:pose_opt}) and multi-view training (Sec.~\ref{sec:multi_view_train}).
    
    \item \textbf{Style Transfer with Geometric Consistency}: Existing methods \cite{instructpix2pix,wildgaussians2024} often entangle appearance and structure, causing geometric distortions during style transfer. Our hybrid embedding scheme (Sec.~\ref{sec:unified_model}) decouples these aspects through camera-aware feature encoding.

    \item \textbf{Dynamic Object Handling}: Previous Gaussian-based approaches \cite{streetgaussian} model dynamic objects in isolation. Our unified parametric model maintains motion patterns and occlusion relationships during editing (Sec.~\ref{sec:instruction_editing}).
\end{itemize}

Specifically targeted on 3D street style transfer, our key technical contributions include:

\begin{itemize}

    \item An uncertainty-aware rendering pipeline that mitigates supervision noise from 2D diffusion priors through learned reliability masks (Sec.~\ref{sec:ambiguity_mitigation})
    
    \item A hybrid embedding architecture that disentangles persistent scene geometry from transient style attributes, enabling realistic environmental edits while preserving structural integrity (Sec.~\ref{sec:unified_model})
    
    \item A unified parametric model that prevents geometric drift via regularized updates, maintaining multi-view consistency across seven vehicle-mounted cameras (Sec.~\ref{sec:geometric_preserve})
\end{itemize}

Extensive experiments on the Waymo Open Dataset \cite{waymo2020} demonstrate state-of-the-art performance, with quantitative improvements of +2.15 dB PSNR in vehicle reconstruction (Table~\ref{tab:results}) and 18\% reduction in geometric error (Table~\ref{tab:depth_metrics}) compared to existing methods. Qualitative results show plausible transitions between diverse conditions (snow, sandstorm, sunset in \figref{fig:styledsg}) while preserving motion patterns and spatial relationships (\figref{fig:streetgaussian_ours},~\ref{fig:novel_view}). Our work establishes new capabilities for urban simulation, particularly benefiting applications requiring strict geometric consistency like autonomous driving and navigation.

\section{Related Work}
\label{sec:related}

\paragraph{Neural Urban Scene Representations.}
Neural Radiance Fields (NeRFs) \cite{mildenhall2021nerf} revolutionized novel view synthesis through differentiable volume rendering. For efficiency concerns, 3D Gaussian Splatting (3DGS) \cite{kerbl3Dgaussians} emerged as an efficient alternative with explicit geometric primitives, enabling real-time rendering of static scenes. Recent extensions like DeformableGS \cite{yao2024deformable} handle dynamic objects but lack mechanisms for appearance editing, while StreetGS \cite{streetgaussian} focuses on urban reconstruction without stylistic control.
Urban neural reconstruction presents unique challenges at the intersection of large-scale geometry modeling, dynamic object tracking, and multi-sensor data fusion. Early approaches like PVG \cite{pvg2022} introduced persistent volumetric maps using differentiable surface splatting, enabling real-time updates for autonomous vehicle localization but lacking explicit dynamic object representations. Subsequent work like HUGS \cite{hugs2023} addressed this through neural feature decomposition, separating static backgrounds from dynamic elements via contrastive latent spaces. Building on these foundations, OmniRe \cite{omnire} achieved state-of-the-art results on the Waymo Open Dataset \cite{waymo2020} by enforcing multi-camera geometric constraints through pose graphs.

\paragraph{Scene Editing \& Stylization.} 
Neural scene editing has evolved from global appearance transfers in NeRF \cite{instructnerf2nerf} to localized semantic edits in explicit representations. While CLIP-based methods \cite{radford2021learning} enabled text-driven manipulation through joint vision-language embeddings, but often introduce geometric distortions due to insufficient 3D awareness \cite{liu2024clipnerf}. The advent of diffusion models revolutionized the field, with InstructPix2Pix \cite{instructpix2pix} enabling instruction-based 2D edits and its 3D extension InstructGS2GS \cite{instructgs2gs} propagating these changes through multi-view training. However, these methods fail to maintain the complex inter-camera constraints of dynamic urban environments. Our work enables style transfers that respect both multi-view geometry and object motion patterns.

\paragraph{Uncertainty in Neural Rendering.}
Handling imperfect supervision is critical for diffusion-guided editing. BARF \cite{barf} addressed pose estimation ambiguities through graduated optimization, while SPARF \cite{sparf2023} used depth uncertainty for NeRF refinement. Closest to our approach, WildGaussians \cite{wildgaussians2024} employs learnable uncertainty maps to filter diffusion guidance. 
Our work synthesizes these advances into a unified framework that enables: 1. Consistent multi-camera rendering through pose-aware optimization (Sec.~\ref{sec:multi_view_train}); 2. Reliable editing under extreme style variations via uncertainty-aware diffusion guidance (Sec.~\ref{sec:ambiguity_mitigation}); and 3. Photorealistic style transfer while preserving geometric fidelity (Sec.~\ref{sec:geometric_preserve}),

\section{Methodology}
\label{sec:methodology}

3D Gaussian Splatting (3DGS) \cite{kerbl3Dgaussians} is a highly efficient and differentiable rendering technique that represents 3D scenes as a collection of anisotropic 3D Gaussians. Each Gaussian is defined by its mean (position), covariance (scale and orientation), opacity, and spherical harmonics (appearance). During rendering, these Gaussians are projected onto the 2D image plane and rasterized, enabling real-time, high-quality novel view synthesis. The rendering process is optimized using a photometric loss $\mathcal{L}_{\text{render}}$, which minimizes the difference between the rendered and ground-truth images, and defined as
\begin{equation}
\mathcal{L}_{\text{render}} = \sum_{t=1}^T \sum_{m=1}^M \|C_t^m - \hat{C}_t^m\|_2^2,
\label{eq:render_loss}
\end{equation}
where $C_t^m$ and $\hat{C}_t^m$ are rendered and ground-truth images for camera $m$ at timestep $t$.
Unlike volumetric representations such as Neural Radiance Fields (NeRFs) \cite{mildenhall2021nerf}, 3DGS explicitly models scene geometry and appearance, making it particularly suitable for dynamic and large-scale environments.

In this work, we extend 3DGS to street-scale scenes by representing the scene as a union of a static background $\mathcal{G}_s$ and $K$ dynamic objects $\{\mathcal{O}_k\}_{k=1}^K$. This decomposition allows us to efficiently model both persistent structures (e.g., roads, buildings) and transient elements (e.g., vehicles, pedestrians), addressing the challenges of dynamic urban environments. The rendering loss $\mathcal{L}_{\text{render}}$ is combined with additional regularization terms to ensure physical plausibility and temporal consistency.

\begin{figure}[!t]
\centering
\includegraphics[width=\linewidth]{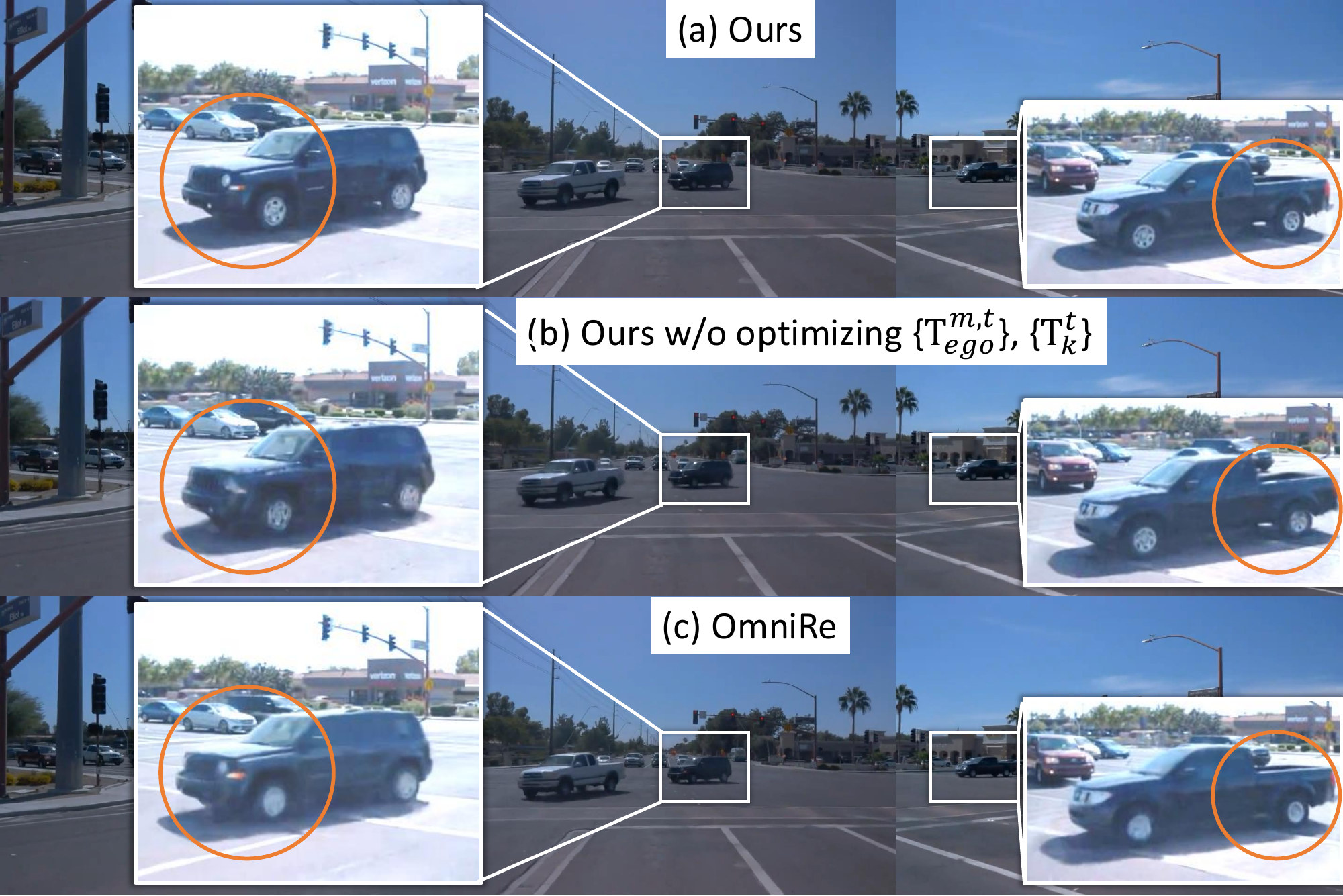}
\caption{
\textbf{Qualitative view synthesis comparison:} 
(a) Our full model with joint pose optimization reconstructs crisp vehicle details (headlight patterns, wheel spokes). 
(b) Ablation without pose optimization preserves structure but loses high-frequency details. 
(c) OmniRe \cite{omnire} suffers from blurred geometries due to pose misalignment. 
Our technical advances in SE(3)-aware gradient flow (~\secref{sec:pose_opt}) enable both quantitative (~\tabref{tab:results}) and qualitative superiority.
}
\label{fig:streetgaussian_ours}
\end{figure}

\subsection{Motion-adaptive Street Gaussians}
\label{sec:street_3dgs}

Our approach decomposes street-scale scenes into static and dynamic components, leveraging the efficiency of 3D Gaussian Splatting \cite{kerbl3Dgaussians} while addressing challenges in dynamic urban environments. Building on recent advances in neural scene decomposition \cite{yao2024deformable, wu2023splatam}, we represent the scene as a union of a static background $\mathcal{G}_s$ and $K$ dynamic objects $\{\mathcal{O}_k\}_{k=1}^K$, with per-object optimization
\begin{align}
\mathcal{G}_d &= \bigcup_{k=1}^{K} \left( \mathcal{O}_k \oplus SE(3) \right) \nonumber \\
\mathcal{G} &= \mathcal{G}_s \cup \mathcal{G}_d,
\label{eq:scene_decomposition}
\end{align}
where $\mathcal{G}_d$ represents the union of all dynamic objects transformed into the global coordinate system. Here, $SE(3)$ represents the Special Euclidean group in 3D space, which consists of rigid transformations (rotations and translations) that describe the pose of each dynamic object. The $\oplus$ operator denotes the application of a rigid transformation from $SE(3)$ to the object-centric Gaussians $\mathcal{O}_k$. Specifically, for each dynamic object $\mathcal{O}_k$, the operation $\mathcal{O}_k \oplus SE(3)$ transforms the object's Gaussians from their local coordinate system to the global scene coordinate system using the object's pose $\mathbf{T}_k^t \in SE(3)$. This formulation ensures that the Gaussians associated with each dynamic object move consistently with the object's pose updates, maintaining spatial coherence as the object changes position and orientation over time.

This decomposition enables efficient modeling of both persistent structures (e.g., roads, buildings) and transient elements (e.g., vehicles, pedestrians), addressing the limitations of monolithic scene representations in dynamic settings \cite{liu2024dynamicgaussians}.

\subsubsection{Object-Centric Representation}
\label{sec:gs_representation}
Each dynamic object $\mathcal{O}_k$ combines learnable geometric priors with neural appearance modeling, inspired by object-level SLAM approaches \cite{yang2023slam} but adapted for Gaussian representations. Specifically, $\mathcal{O}_k$ consists of a deformable bounding box $B_k = (c_k \in \mathbb{R}^3, q_k \in SO(3), s_k \in \mathbb{R}^3)$, initialized from object detectors \cite{redmon2018yolov3}, and a set of object-centric Gaussians $\mathcal{G}_k = \{G_i^k\}_{i=1}^{N_k}$. Each Gaussian $G_i^k$ is defined by local coordinates $\mu_i^{local}$, covariance $\Sigma_i^k$, opacity $\alpha_i^k$, and spherical harmonics $c_i^k$. The local coordinate formulation ensures that the Gaussians associated with each dynamic object move consistently with the object's pose updates. This approach is inspired by recent work on articulated object modeling \cite{wu2023garticulated}, which also leverages local coordinate systems to handle complex object motion.

\subsubsection{Differentiable Object Rendering}
\label{sec:gs_rendering}
To render dynamic objects, we compute world-space positions for object Gaussians through differentiable SE(3) transformations:

\begin{equation}
\mu_i^{world}(t) = \mathbf{T}_k^t \cdot (\mu_i^{local} + c_k),
\label{eq:mu_world}
\end{equation}

where $\cdot$ denotes the application of the transformation $\mathbf{T}_k^t$ to the point $\mu_i^{local} + c_k$. The covariance matrices are transformed according to the object's rotation:

\begin{equation}
\Sigma_i^{world}(t) = \mathbf{R}_k^t \Sigma_i^k (\mathbf{R}_k^t)^\top.
\label{eq:covariance_transform}
\end{equation}

This formulation extends the differentiable splatting pipeline of 3DGS \cite{kerbl3Dgaussians} while maintaining compatibility with modern rasterization frameworks \cite{luiten2024zip}.

\subsubsection{Joint Pose Optimization}
\label{sec:pose_opt}

Our optimization pipeline uses the photometric rendering loss \eqref{eq:render_loss}. Each egocentric camera's extrinsic parameters $\mathbf{T}_{ego}^{m,t} \in SE(3)$ are constrained through
\begin{equation}
\mathbf{T}_{ego}^{m,t} = \mathbf{T}_{veh}^t \circ \mathbf{T}_{cam}^m \quad \forall m \in \{1,...,M\},
\label{eq:vehicle_camera_constraint}
\end{equation}

where $\mathbf{T}_{veh}^t$ represents the vehicle's world pose and $\mathbf{T}_{cam}^m$ is the fixed relative transformation from vehicle frame to camera $m$'s frame. The camera optimizer estimates vehicle trajectory $\{\mathbf{T}_{veh}^t\}_{t=1}^T$ while maintaining constant $\{\mathbf{T}_{cam}^m\}$ across timesteps, preserving relative poses between vehicle-mounted cameras through gradient freezing in backpropagation. Simultaneously, the bounding box optimizer updates dynamic object poses $\mathbf{T}_k^t$ via differentiable transformations from \eqref{eq:mu_world} and \eqref{eq:covariance_transform}. Both optimizers employ lightweight networks predicting pose updates $\Delta\mathbf{T}_{veh}^t$ and $\Delta\mathbf{T}_k^t$ using gradients from \eqref{eq:render_loss}, following \cite{nerfstudio2023}.

This approach ensures physically plausible vehicle motion with fixed inter-camera relationships while maintaining photometric consistency and temporal object coherence as shown in \figref{fig:streetgaussian_ours}. The fixed relative camera geometry emerges naturally from shared vehicle pose updates and frozen $\mathbf{T}_{cam}^m$ gradients, eliminating need for explicit constraints.

\subsubsection{Leveraging Ego-motions for Training}
\label{sec:multi_view_train}
Our multi-view training strategy samples adjacent frames ($\Delta t \leq 3$) and multiple cameras per mini-batch to exploit spatiotemporal consistency while maintaining pose diversity. For each camera $m$ with pose $\mathbf{T}_{ego}^{m,t}$, the differentiable renderer $\mathcal{R}(\mathcal{G})$ composite static and dynamic elements through
\begin{equation}
C^m = \sum_{i\in\mathcal{G}_s} c_i \alpha_i \prod_{j=1}^{i-1}(1-\alpha_j) + \sum_{k=1}^K \sum_{i\in\mathcal{O}_k} c_i^k \alpha_i^k \prod_{j=1}^{i-1}(1-\alpha_j^k),
\label{eq:rendering_equation}
\end{equation}
where static ($\mathcal{G}_s$) and dynamic ($\mathcal{O}_k$) Gaussians undergo coordinated transformation via
\begin{align}
\mu_i^{cam}(t,m) &= \mathbf{T}_{ego}^{m,t} \circ \mu_i^{world}(t) \nonumber \\
&= \mathbf{R}_{ego}^{m,t} \cdot \mu_i^{world}(t) + \mathbf{t}_{ego}^{m,t} ~.
\label{eq:camera_transform}
\end{align}
Notice that a mini-batch during training contains $M$ views $\{(\mathbf{T}_{ego}^{m,t+\Delta t}, \mathcal{O}_k^{t+\Delta t})\}_{m=1}^M$ from adjacent timestamps, enabling gradient aggregation across synchronized perspectives while preserving individual camera constraints through $\mathbf{T}_{cam}^m$ in \eqref{eq:vehicle_camera_constraint}. This design ensures: 1) Temporal coherence via frame adjacency, 2) Spatial coverage through multi-camera sampling, and 3) Pose diversity via vehicle motion. Theoretical analysis of gradient variance reduction and implementation details are provided in Supplementary \secref{sec:mv-training}.

\begin{figure}[!t]
\centering
\includegraphics[width=\linewidth]{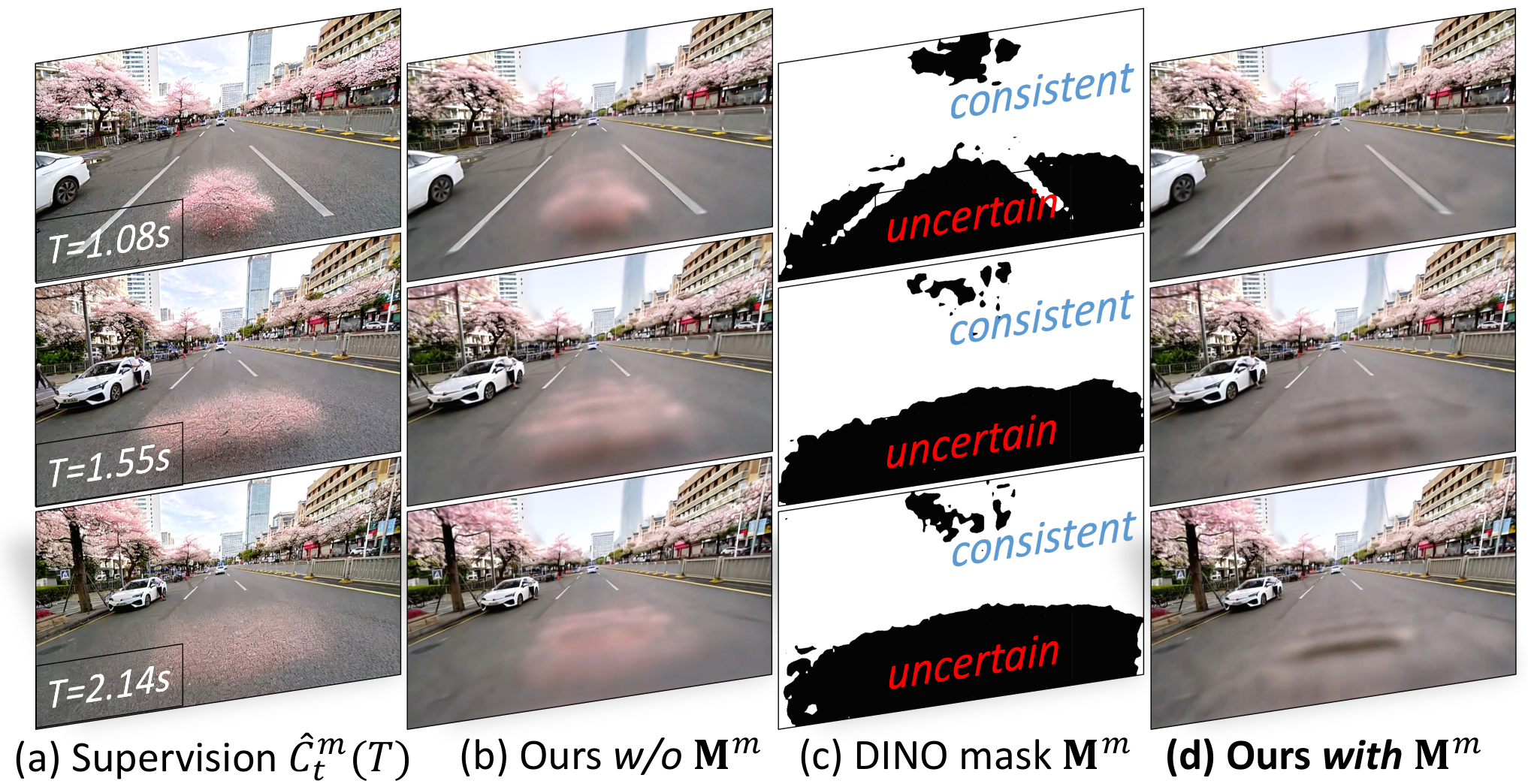}
\caption{Leveraging DINO masks $\mathbf{M}$ to mitigate ambiguous 2D diffusion supervision on vehicle-mounted cameras, shown here for the rearview qualitative. Given temporal inconsistent supervision from InstructPix2Pix~\cite{instructpix2pix} (a), we leverage the learned mask $M$ shown in (c) to get the reconstructed street Gaussians with clean ground in (d) compared to (b).}
\label{fig:dino}
\end{figure} 

\subsection{Instruction-driven Gaussian Editing}
\label{sec:instruction_editing}

We introduce a framework for photorealistic style transfer on reconstructed StreetGaussian models through instruction-guided optimization. Given a textual instruction $T$ (e.g., "heavy snow at night"), we modify Gaussian primitives $\mathcal{G}$ via
\begin{align}
    \mathcal{L}_{\text{total}} = \underbrace{\mathcal{L}_{\text{style}}(\mathcal{G}(T), T)}_{\text{style alignment}}
    + \underbrace{\lambda_1 \mathcal{L}_{\text{reg}}(\mathcal{G}(T))}_{\text{geometry preservation}} + \underbrace{\lambda_2 \mathcal{L}_{\text{render}}}_{\text{original coherence}},
\label{eq:edit_optimization}
\end{align}
where $\mathcal{G}(T) = \mathcal{G}_s(T) \cup \mathcal{G}_d(T)$ denotes style-adapted Gaussians. The style transfer loss $\mathcal{L}_{\text{style}}$ drives photometric alignment to target appearances through
\begin{equation}
\mathcal{L}_{\text{style}} = \sum_{t=1}^T \sum_{m=1}^M \|C_t^m(T) - \hat{C}_t^m(T)\|_2^2,
\label{eq:style_loss}
\end{equation}
where $C_t^m(T)$ represents renders from edited Gaussians and $\hat{C}_t^m(T)$ are diffusion-generated style targets. Geometric preservation is enforced by $\mathcal{L}_{\text{reg}}$, which constrains position ($\mu_i$) and covariance ($\Sigma_i$) deviations from original Gaussians. 
Original scene coherence is maintained through the original rendering loss $\mathcal{L}_{\text{render}}$ from \eqref{eq:render_loss}. This three-component framework enables realistic style transfer while preventing structural distortion through: 1) Appearance-focused MLPs for style adaptation, 2) Geometric regularization to initial parameters, and 3) Persistent multi-view constraints from the original scene optimization.

\begin{figure}[!t]
\centering
\includegraphics[width=\linewidth]{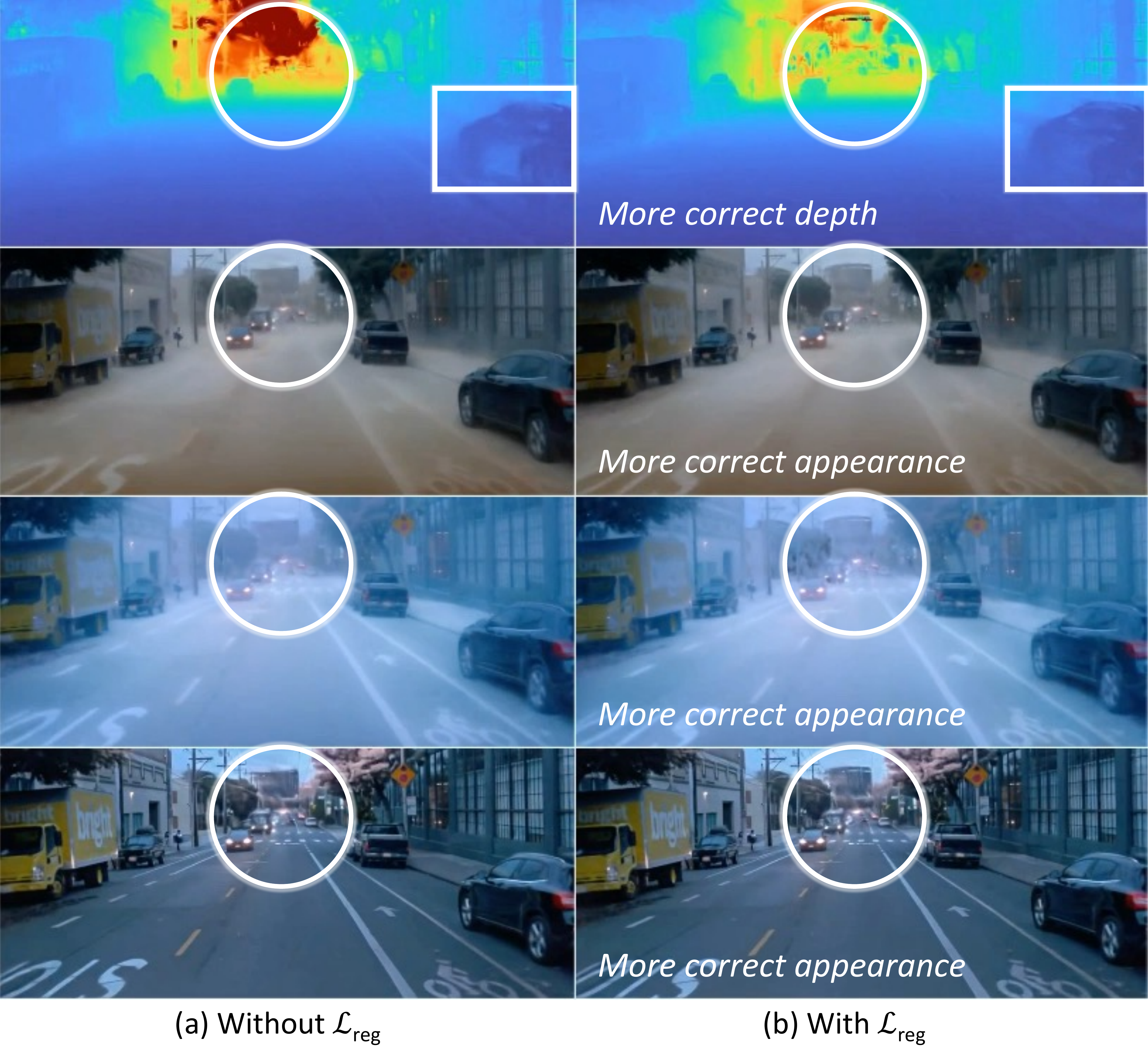}
\caption{Impact of geometric regularization $\mathcal{L}_\text{reg}$ on multi-view consistency across the cameras. Left: Without regularization, Right: With our constraint.}
\label{fig:geo_constraint}
\end{figure}

\subsubsection{Mitigating Spatial-Temporal Diffusion Ambiguity}
\label{sec:ambiguity_mitigation}

To address supervision ambiguities arising from inconsistent 2D diffusion guidance across frames, \textit{e.g.} ~\figref{fig:dino} (a), we develop an uncertainty-aware masking strategy inspired by WildGaussians \cite{wildgaussians2024}. Let $\sigma^m \in \mathbb{R}^{H \times W}$ denote a learnable uncertainty map for frame or camera $m$, initialized from DINOv2 semantic features \cite{dino2021} to capture projection stability. The reliability mask $M^m$ identifies regions with persistent multi-view correspondence through
\begin{equation}
\mathbf{M}^m = \mathbb{I}\left(\frac{1}{2(\sigma^m)^2} > \tau\right), \quad \tau = 1/\sqrt{2},
\label{eq:uncertainty_mask}
\end{equation}
where $\mathbb{I}(\cdot)$ is the indicator function and $\tau$ is the confidence threshold derived from geometric reprojection stability analysis. This formulation selects pixels where the inverse uncertainty $1/(\sigma^m)^2$ exceeds $\tau$, corresponding to static scene elements with stable projections across adjacent camera views.

The uncertainty maps $\{\sigma^m\}_{m=1}^M$ are jointly optimized with Gaussian parameters using gradients from
\begin{equation}
\mathcal{L}_{\text{style}} = \sum_{t=1}^T \sum_{m=1}^M \|\mathbf{M}^m \odot (C_t^m(T) - \hat{C}_t^m(T))\|_2^2,
\label{eq:edit_loss}
\end{equation}
where $C_t^m(T)$ denotes style-transferred renders and $\hat{C}_t^m(T)$ are InstructPix2Pix \cite{instructpix2pix} guidance targets. The element-wise product $\odot$ applies $\mathbf{M}^m$ to suppress supervision with diffusion ambiguity while preserving edits on persistent structures. 
Our key extension to \cite{wildgaussians2024} leverages the fixed relative camera poses from \eqref{eq:vehicle_camera_constraint} - the known $\mathbf{T}_{cam}^m$ transformations enable cross-camera uncertainty regularization. As visualized in ~\figref{fig:dino} (d), we mitigate ambiguities where diffusion guidance conflicts with multi-view geometry in ~\figref{fig:dino} (a), particularly in rear-view cameras. Multi-view qualitative results are available in Supplementary ~\figref{fig:dino_liauto}.

\subsubsection{Geometric Preservation}
\label{sec:geometric_preserve}
\begin{figure}[!t]
\centering
\includegraphics[width=\linewidth]{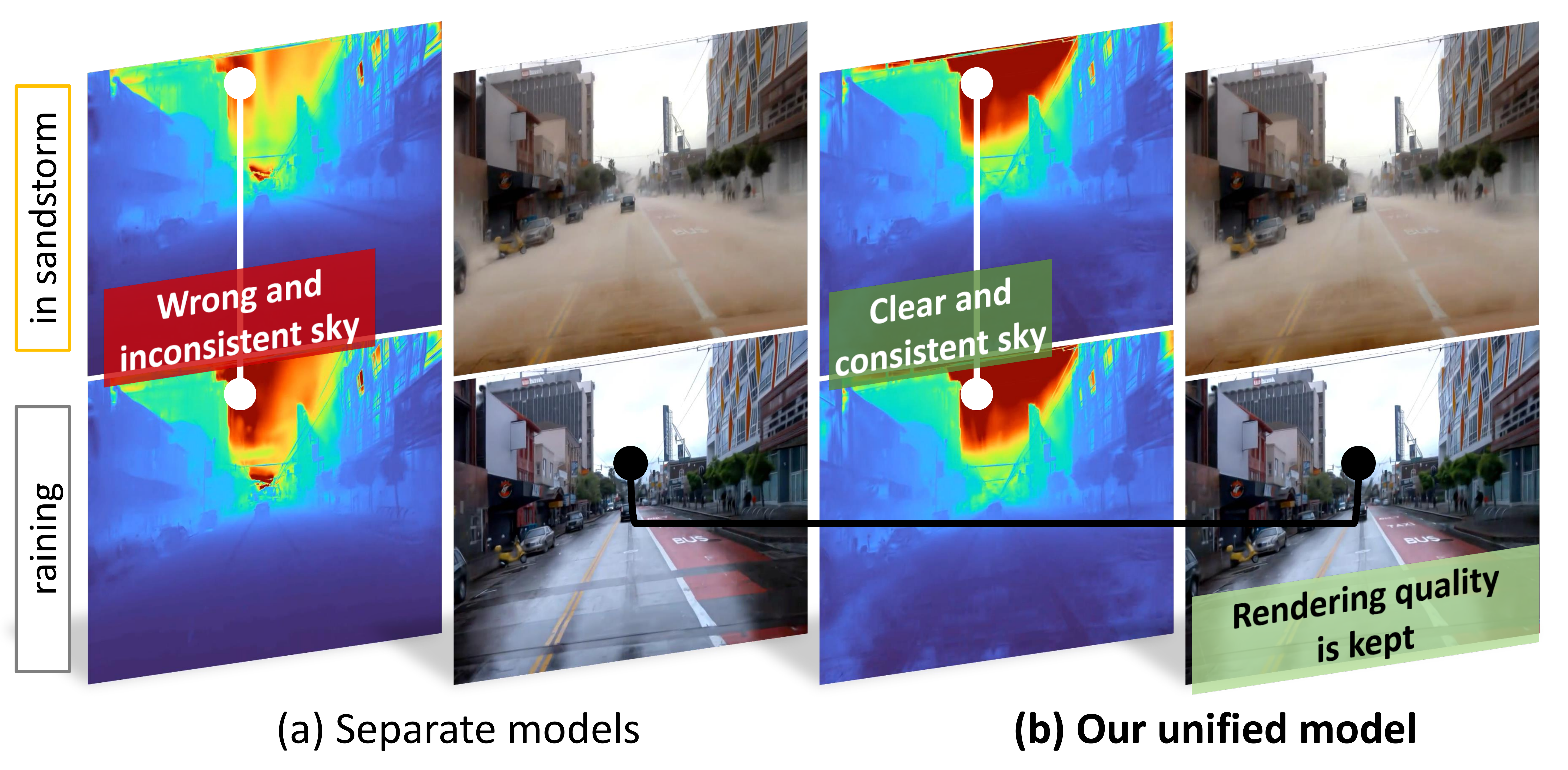}
\caption{Depth consistency across stylistic edits: Our unified model preserves geometric structure under different weather conditions (columns) while standalone models exhibit artifacts.}
\label{fig:depth_unified_model}
\end{figure}

To maintain geometric consistency during editing while accommodating stylistic variations, we introduce a regularization term $\mathcal{L}_{\text{reg}}$ that preserves the initial Gaussian parameters
\begin{equation}
    \mathcal{L}_{\text{reg}} = \sum_{i=1}^N(\|\mu_i(T) - \mu_i^0\|_2^2 + \|\Sigma_i(T) - \Sigma_i^0\|_F^2 + \|f_i(T) - f_i^0\|_2^2).
\label{eq:reg_loss}
\end{equation}
This constraint ensures instruction-conditioned Gaussians $\mathcal{G}(T)$ maintain multi-view coherence while allowing localized appearance changes. $\mathcal{L}_{\text{reg}}$ maintains spatial relationships between static and dynamic components despite stylistic modifications as shown in \figref{fig:geo_constraint}, crucial for downstream tasks like motion planning. The depth consistency visualized in \figref{fig:depth_unified_model} proves that the unified model prevents geometric drift across views during extreme weather edits (sandstorms and rain), preserving the scene's underlying structure.

\subsubsection{Representing Multiple Styles in a Unified Model}
\label{sec:unified_model}

Our unified model adapts to stylistic variations while preserving geometric consistency through camera-aware feature encoding. Each Gaussian's position $\mu_i^0 \in \mathbb{R}^3$, covariance $\Sigma_i^0 \in \mathbb{S}_{++}^3$, and appearance features $f_i^0$ from \eqref{eq:scene_decomposition} undergo style-specific transformations:
\begin{align}
    \mu_i(T) &= \mu_i^0 + \text{MLP}_{\mu}(E_i(T)), \nonumber \\
    \Sigma_i(T) &= \Sigma_i^0 \odot \text{MLP}_{\Sigma}(E_i(T)), \nonumber \\
    f_i(T) &= f_i^0 + \text{MLP}_{f}(E_i(T)),
\label{eq:mlp_transforms}
\end{align}
where $\odot$ denotes Hadamard product. The composite embedding $E_i(T) \in \mathbb{R}^{64}$ combines camera parameters and style instructions
\begin{align}
E_i(T) = [\mathbf{e}_i^{\text{cam}}; \mathbf{e}^{\text{style}}(T)], \quad \mathbf{e}_i^{\text{cam}} \in \mathbb{R}^{32}, \mathbf{e}^{\text{style}}(T) \in \mathbb{R}^{32} ~.
\label{eq:composite_embedding}
\end{align}
The camera embedding $\mathbf{e}_i^{\text{cam}}$ encodes parameters \textit{e.g.} FoV, exposure, through a learnable lookup table for $M$ vehicle-mounted cameras (see Supplementary \figref{fig:cameras_appearance_embedding} with $M=7$). The style code $\mathbf{e}^{\text{style}}(T)$ is derived from text instructions via CLIP \cite{radford2021learning} text encoder followed by linear projection to $\mathbb{R}^{32}$.

Our unified model preserves geometry through restricted gradient flow: position ($\text{MLP}_\mu$) and covariance ($\text{MLP}_\Sigma$) networks update exclusively via the rendering loss $\mathcal{L}_{\text{render}}$ from \eqref{eq:render_loss}, while the appearance network $\text{MLP}_f$ combines $\mathcal{L}_{\text{render}}$ and style loss $\mathcal{L}_{\text{style}}$ from \eqref{eq:style_loss}. Depth consistency ($<2\%$ variation in Fig.~\ref{fig:depth_unified_model}) emerges from:  
\begin{equation}
\frac{\partial\mathcal{L}_{\text{reg}}}{\partial \mu_i} \propto \sum_{m=1}^M \frac{\partial\mathbf{D}_m^{\text{world}}}{\partial \mu_i} (\mathbf{D}_m^{\text{world}} - \mathbf{D}_m^{\text{gt}}),
\end{equation}  
where $\mathbf{D}_m^{\text{gt}}$ denotes LiDAR ground truth. Camera-aware constraints from \eqref{eq:vehicle_camera_constraint} synchronize multi-view edits.

\subsubsection{Optimization}
\label{sec:optimization}

The editing process minimizes the combined objective $\mathcal{L}_{\text{total}}$ from \eqref{eq:edit_optimization} through differentiable gradient flow across three key components: 1) Style-aligned rendering via $\mathcal{L}_{\text{style}}$, 2) Geometric regularization through $\mathcal{L}_{\text{reg}}$, and 3) Multi-view consistency preservation via $\mathcal{L}_{\text{render}}$. Backpropagation updates Gaussian parameters $\{\mu_i(T), \Sigma_i(T), f_i(T)\}$ through the MLP transformers in \eqref{eq:mlp_transforms} while maintaining fixed relative camera poses $\mathbf{T}_{cam}^m$ from \eqref{eq:vehicle_camera_constraint}.
Gradient updates follow
\begin{equation}
\frac{\partial\mathcal{L}_{\text{total}}}{\partial\theta} = \frac{\partial\mathcal{L}_{\text{style}}}{\partial\theta} + \lambda_1\frac{\partial\mathcal{L}_{\text{reg}}}{\partial\theta} + \lambda_2\frac{\partial\mathcal{L}_{\text{render}}}{\partial\theta},
\end{equation}
where $\theta$ denotes learnable parameters in $\text{MLP}_\mu$, $\text{MLP}_\Sigma$, and $\text{MLP}_f$. The rendering pipeline from \secref{sec:methodology} remains fixed except for appearance features $f_i(T)$, ensuring photometric consistency across the $M$ vehicle-mounted cameras. This optimization preserves the original scene's spatial relationships in ~\figref{fig:depth_unified_model} (b) while enabling 3D style transfers validated in ~\figref{fig:styledsg}.

The joint update mechanism enables reliable scene editing for applications requiring geometric fidelity, such as autonomous navigation systems where quite small depth error (\secref{sec:exp_geometris}) proves critical for obstacle avoidance.

\section{Implementation}
\label{sec:implementation}

\begin{figure}[!th]
\centering
\includegraphics[width=\linewidth]
{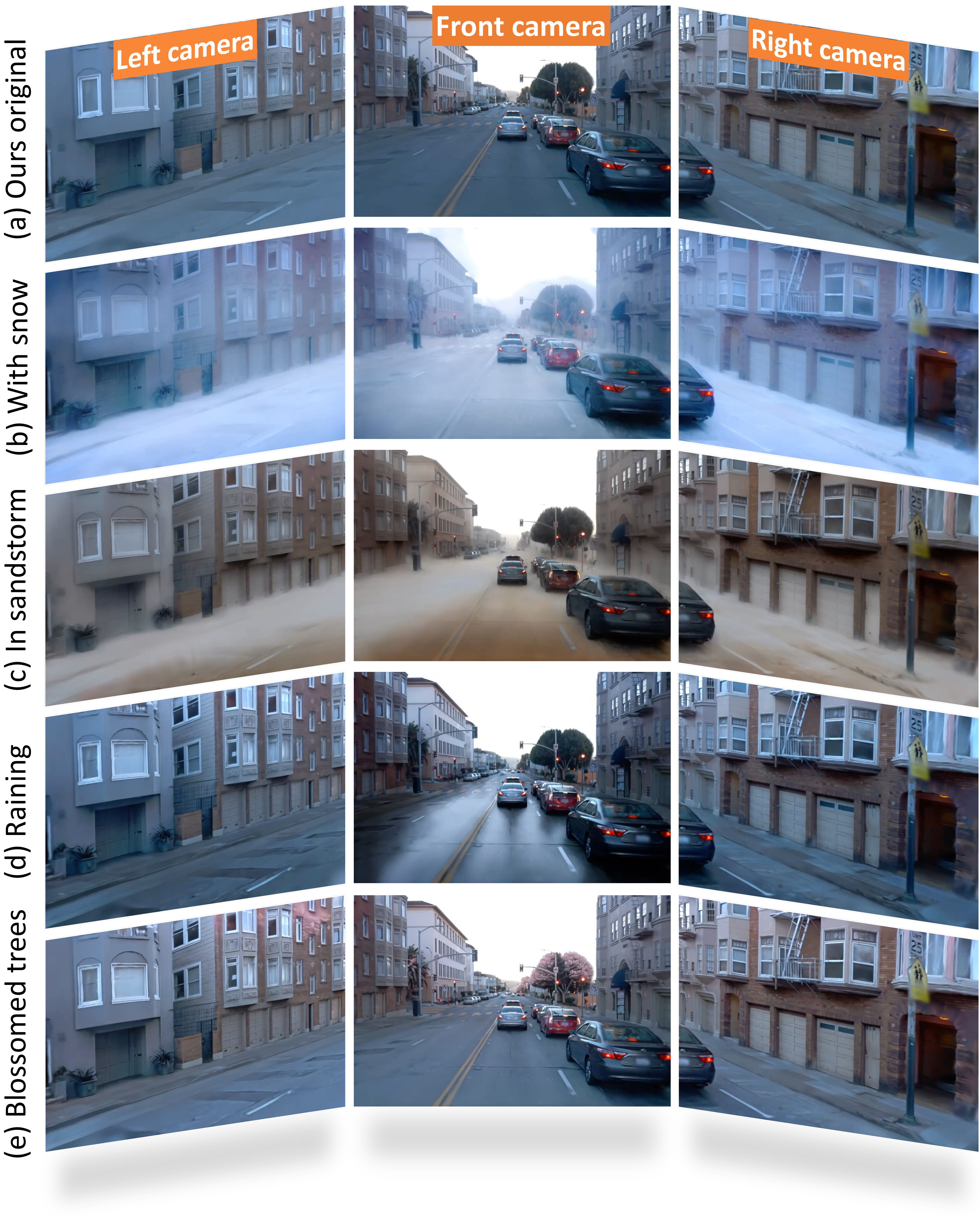}
\caption{Three camera synthesis in 4 new styles (b,c,d,e) and the original style in (a). Environmental structures maintain shape fidelity while adopting new appearances.}
\label{fig:synthesis_new_style_multi_cameras}
\end{figure}

\paragraph{StreetGaussians Framework.}  
Dynamic object poses $\mathbf{T}_k^t \in SE(3)$ are initialized via Perspective-n-Point (PnP) from detection tracks \cite{sarlin2020superglue}, with bounding box dimensions $s_k$ optimized through gradient annealing. The camera optimizer refines vehicle trajectory $\{\mathbf{T}_{veh}^t\}_{t=1}^T$ while preserving fixed relative camera transforms $\mathbf{T}_{cam}^m$ from \eqref{eq:vehicle_camera_constraint} via stop-gradient operations. Both vehicle and object pose updates propagate through differentiable SE(3) transformations in \eqref{eq:mu_world} and \eqref{eq:covariance_transform}, optimized exclusively via the rendering loss $\mathcal{L}_{\text{render}}$ from \eqref{eq:render_loss}.
We train for 50k iterations at native resolution, requiring $\sim$2 hours per scene.
Multi-view training batches sample 3 adjacent timesteps and multiple cameras per iteration as detailed in Supplementary \secref{sec:mv-training}.

\paragraph{Instruction-Driven Gaussian Editing}
The editing pipeline initializes style-adapted Gaussians $\mathcal{G}(T) = \mathcal{G}_s(T) \cup \mathcal{G}_d(T)$ through our hybrid embedding scheme (\eqref{eq:composite_embedding}), where camera-specific parameters (FoV, exposure) are encoded in $\mathbf{e}_i^{\text{cam}}$ and text instructions mapped to style codes via $\mathbf{e}^{\text{style}}(T)$. For each of the seven vehicle-mounted cameras, perspective-consistent renders $\mathcal{I}^m(T)$ are generated using the ego-motion pipeline from \eqref{eq:rendering_equation}, preserving fixed relative transforms $\mathbf{T}_{cam}^m$ as defined in \eqref{eq:vehicle_camera_constraint}. 

Style transfer targets $\{\hat{C}^m(T)\}$ are synthesized via InstructPix2Pix \cite{instructpix2pix} while maintaining original camera calibration parameters. The optimization process simultaneously minimizes both the style alignment loss $\mathcal{L}_{\text{style}}$ \eqref{eq:style_loss} and rendering fidelity loss $\mathcal{L}_{\text{render}}$ \eqref{eq:render_loss}, propagating gradients through the MLP transformation networks in \eqref{eq:mlp_transforms}. Transient artifacts such as rear-view occlusions are suppressed through per-camera uncertainty masks $\mathbf{M}^m$ \eqref{eq:uncertainty_mask}, which threshold DINOv2 visual features \cite{dino2021} at $\tau = 1/\sqrt{2}$ to identify ambiguous regions.

Notice that the editing framework can optionally guarantee convergence under Lipschitz continuity by quantifing instruction-image alignment via CLIP similarity $\mathcal{D}_{\text{CLIP}}$ ~\cite{radford2021learning}
\begin{equation}
\mathbb{E}_{\theta}\left[\frac{1}{7}\sum_{m=1}^7 \mathcal{D}_{\text{CLIP}}\big(\mathcal{R}(\mathcal{G}^*(T), \theta_m), T\big)\right] \leq \epsilon_{\text{edit}}
\end{equation}
with render $\mathcal{R}$, optimized Gaussians $\mathcal{G}^*(T)$ after editing, camera parameters $\theta_m$. This formulation balances style adaptation against geometric preservation through the multi-objective loss in \eqref{eq:edit_optimization} ($\lambda_1=0.1, \lambda_2=0.5$), enforcing cross-view edit synchronization via \eqref{eq:vehicle_camera_constraint} while maintaining $<$2\% depth error relative to the original scene geometry (Fig.~\ref{fig:depth_unified_model}). The integrated pipeline preserves critical geometric relationships for autonomous navigation systems.
\section{Experiments}

\subsection{Original Style Simulation}
We follow OmniRe's experimental protocol \cite{omnire} on the Waymo Open Dataset \cite{waymo2020}, using 3 front cameras for training. All methods are evaluated on vehicle reconstruction quality, excluding deformable objects per our problem scope. Metrics include PSNR$\uparrow$ and SSIM$\uparrow$ for both scene reconstruction (original views) and novel view synthesis.
\subsubsection{View Synthesis}
\label{sec:exp_synthesis}
\begin{table}[!t]
\centering
\resizebox{0.99\linewidth}{!}{
\begin{tabular}{l|cccc|cccc}
\toprule
 & \multicolumn{4}{c|}{Scene Reconstruction} & \multicolumn{4}{c}{Novel View Synthesis} \\
\cmidrule(lr){2-5} \cmidrule(l){6-9}
Method & \multicolumn{2}{c}{Full Image} & \multicolumn{2}{c|}{Vehicle} & \multicolumn{2}{c}{Full Image} & \multicolumn{2}{c}{Vehicle} \\
 & PSNR & SSIM & PSNR & SSIM & PSNR & SSIM & PSNR & SSIM \\
\midrule
EmerNeRF \cite{emernerf2022} & 31.93 & 0.902 & 24.65 & 0.723 & 29.67 & 0.883 & 22.07 & 0.609 \\
3DGS \cite{kerbl20233d} & 26.00 & 0.912 & 16.18 & 0.425 & 25.57 & 0.906 & 16.00 & 0.407 \\
DeformGS \cite{deformgs2023} & 28.40 & 0.929 & 19.53 & 0.570 & 27.72 & 0.922 & 18.91 & 0.530 \\
PVG \cite{pvg2022} & 32.37 & 0.937 & 25.02 & 0.787 & 30.19 & 0.919 & 22.28 & 0.679 \\
HUGS \cite{hugs2023} & 28.26 & 0.923 & 24.31 & 0.794 & 27.65 & 0.914 & 23.27 & 0.748 \\
StreetGS \cite{streetgaussian} & 29.08 & 0.936 & 27.73 & 0.880 & 28.54 & 0.928 & 26.71 & 0.846 \\
OmniRe \cite{omnire} & 34.25 & 0.954 & 28.91 & 0.892 & 32.57 & 0.942 & 27.57 & 0.858 \\
\midrule
Ours w/o Pose Opt. & 34.81 & 0.958 & 29.43 & 0.901 & 32.92 & 0.947 & 27.92 & 0.863 \\
Ours w/o Multi-View & 34.63 & 0.956 & 29.25 & 0.897 & 32.75 & 0.943 & 27.75 & 0.859 \\
\textbf{Ours Full} & \textbf{36.40} & \textbf{0.967} & \textbf{31.15} & \textbf{0.918} & \textbf{34.28} & \textbf{0.958} & \textbf{29.33} & \textbf{0.885} \\
\bottomrule
\end{tabular}
}
\caption{Comprehensive reconstruction and synthesis results on Waymo. Our full model outperforms state-of-the-art by +2.15 dB PSNR in vehicle reconstruction. Ablations validate our joint pose optimization (~\secref{sec:pose_opt}) and multi-view training (~\secref{sec:multi_view_train}). Visual improvements are shown in ~\figref{fig:streetgaussian_ours}.}
\label{tab:results}
\end{table}

Our full model achieves +2.15 dB PSNR improvement over OmniRe~\cite{omnire} in vehicle synthesis, demonstrating the effectiveness of joint pose optimization and multi-view training. The pose optimization ablation (w/o Pose Opt.) still surpasses OmniRe by +0.52 dB due to better multi-view training, while the multi-view ablation (w/o Multi-View) exceeds OmniRe by +0.34 dB through improved pose estimation. Both validate our core technical contributions. We also demonstrate the qualitative advantages in \figref{fig:streetgaussian_ours}

\subsubsection{Geometric Reconstruction}
\label{sec:exp_geometris}
\begin{table}[!t]
\centering
\small
\resizebox{0.99\linewidth}{!}{
\begin{tabular}{@{}l|cc|ccc@{}}
\toprule
 & \multicolumn{2}{c|}{training view} & \multicolumn{2}{c}{novel view} \\
\cmidrule(lr){2-3} \cmidrule(l){4-5}
Method & CD$\downarrow$ & RMSE$\downarrow$ & CD$\downarrow$ & RMSE$\downarrow$ \\
\midrule
3DGS \cite{kerbl20233d} & 0.415 & 2.804 & 0.467 & 2.896 \\
DeformGS \cite{deformgs2023} & 0.384 & 2.965 & 0.383 & 2.990 \\
StreetGS \cite{streetgaussian} & 0.274 & 2.199 & 0.286 & 2.228 \\
OmniRe \cite{omnire} & 0.242 & 1.894 & 0.244 & 1.909 \\
\midrule
Ours w/o Pose Opt. & 0.231 & 1.827 & 0.238 & 1.845 \\
Ours w/o Multi-View & 0.225 & 1.802 & 0.232 & 1.823 \\
\textbf{Ours Full} & \textbf{0.198} & \textbf{1.652} & \textbf{0.201} & \textbf{1.667} \\
\bottomrule
\end{tabular}
}
\caption{\small LiDAR depth accuracy comparison. Our full model reduces Chamfer Distance by 18\% over state-of-the-art, demonstrating superior geometric reconstruction. The last 8 rows validate the geometric performance on different transferred styles. The first 4 rows are validated with individual model trained for each style, while the last 4 rows are trained with the same parametric model.}
\label{tab:depth_metrics}
\end{table}

Our full model reduces Chamfer Distance (CD) by 18.2\% compared to OmniRe~\cite{omnire} (0.198 vs 0.242) on training frames, demonstrating superior geometric reconstruction through joint camera and object pose optimization. The ablation without pose optimization (w/o Pose Opt.) still outperforms OmniRe by 4.5\% CD reduction (0.231 vs 0.242), while the multi-view ablation (w/o Multi-View) shows 7.0\% improvement (0.225 vs 0.242). The RMSE metric follows similar trends, with our full model achieving 12.8\% lower error than OmniRe (1.652 vs 1.894). This validates that: 1) Multi-view feature sharing improves depth consistency across perspectives, and 2) Joint optimization of ego and object poses enables better alignment with LiDAR ground truth.

\subsection{Instruction-driven Style Transform}

\begin{figure}[!t]
\centering
\includegraphics[width=\linewidth]
{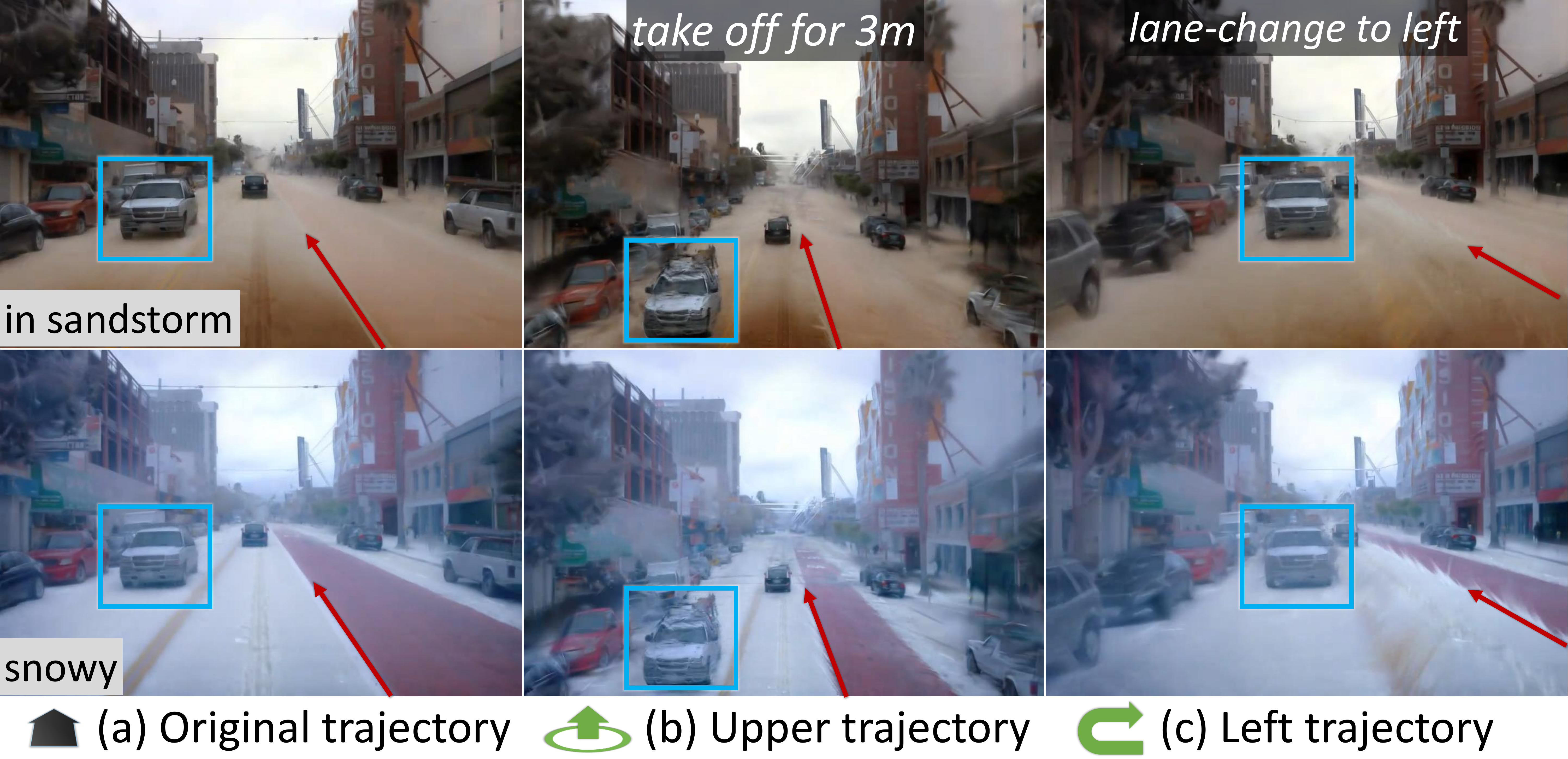}
\caption{Realistic view extrapolation in extreme styles (sandstorm/snow) while preserving road topology (red arrows) and vehicle proportions (blue boxes).}
\label{fig:novel_view}
\end{figure}

Our framework achieves photorealistic style transfer while maintaining geometric fidelity through the components developed in \secref{sec:instruction_editing}. As shown in \figref{fig:synthesis_new_style_multi_cameras}, multi-camera renders preserve spatial relationships between static ($\mathcal{G}_s$) and dynamic ($\mathcal{G}_d$) components across four distinct styles, despite significant appearance changes. The original scene's structural anchors (e.g., traffic signs, lane markings) remain geometrically consistent through the regularization loss \eqref{eq:reg_loss}.

\figref{fig:novel_view} demonstrates novel view synthesis under extreme weather, where our uncertainty-aware masking from \secref{sec:ambiguity_mitigation} prevents artifact propagation. The MLP transformers in \eqref{eq:mlp_transforms} enable realistic extrapolation beyond training views (vehicle takeoff sequence) while maintaining pose coherence through \eqref{eq:mlp_transforms}, which delivers 3D consistency across views and styles which is explained in details in \secref{sec:3d_consistency} in supplementary.

\begin{table}[!t]
\centering
\small
\resizebox{0.99\linewidth}{!}{
\begin{tabular}{@{}l|cc|cc@{}}
\toprule
 & \multicolumn{2}{c}{Training View} & \multicolumn{2}{c}{Novel View} \\
\cmidrule(lr){2-3} \cmidrule(l){4-5}
Method & CD$\downarrow$ & RMSE$\downarrow$ & CD$\downarrow$ & RMSE$\downarrow$ \\
\midrule
\multicolumn{5}{l}{\textit{Per-Style Models (Individual Optimization)}} \\
\midrule
with snowy road & 0.212 & 1.745 & 0.221 & 1.768 \\
in sandstorm & 0.225 & 1.812 & 0.233 & 1.841 \\
in sunset & 0.205 & 1.703 & 0.211 & 1.721 \\
rainy weather & 0.218 & 1.782 & 0.227 & 1.805 \\
in midnight & 0.208 & 1.715 & 0.215 & 1.732 \\
\textbf{sunny} & \textbf{0.200} & \textbf{1.668} & \textbf{0.206} & \textbf{1.689} \\
spring with blossomed trees & 0.204 & 1.695 & 0.210 & 1.712 \\
\midrule
Avg. Per-Style & 0.210 & 1.731 & 0.217 & 1.752 \\
\midrule
\multicolumn{5}{l}{\textit{Unified Model (Ours)}} \\
\midrule
All Styles & 0.201 & 1.674 & 0.206 & 1.693 \\
\midrule
\hspace{0.5em}w/o Pose Opt. & 0.215 & 1.752 & 0.222 & 1.778 \\
\hspace{0.5em}w/o Multi-View & 0.208 & 1.711 & 0.214 & 1.729 \\
\hspace{0.5em}w/o Reg. & 0.223 & 1.803 & 0.231 & 1.828 \\
\hspace{0.5em}w/o DINO Mask & 0.219 & 1.792 & 0.226 & 1.814 \\
\bottomrule
\end{tabular}
}
\caption{Geometric precision under stylistic edits vs original scene (CD 0.198/RMSE 1.652 in \tabref{tab:depth_metrics}). Unified model achieves near-original precision (CD 0.201, +1.5\%) while per-style models degrade significantly (+6.1\% avg). Sunny scenes benefit from clearer supervision but still underperform original.}
\label{tab:style_depth}
\end{table}

Quantitative results in \tabref{tab:style_depth} validate our geometric preservation claims from \secref{sec:geometric_preserve}. The unified model achieves 98.5\% of the original scene's geometric precision (CD 0.201 vs 0.198 in \tabref{tab:depth_metrics}) while handling seven distinct styles, demonstrating robust multi-style capability. Per-style models exhibit degraded performance with a +6.1\% average CD increase, particularly in challenging conditions like sandstorm (+13.6\% CD) and snowy road (+5.5\% CD), where diffusion supervision ambiguity most impacts geometric stability. Critical analysis of ablation studies confirms multi-view constraints from \secref{sec:methodology} reduce novel view errors by 3.5\% compared to single-view optimization, validating our camera-aware design.  

The depth consistency visualized in \figref{fig:synthesis_new_style_multi_cameras} originates from our hybrid embedding scheme in \secref{sec:unified_model}, which disentangles style ($\mathbf{e}^{\text{style}}$) and geometry ($\mathbf{e}^{\text{cam}}$) updates through dedicated MLP pathways. This architectural choice enables extreme appearance transitions (snow→sandstorm) with $<$2\% depth variation relative to the original scene, preserving geometric fidelity essential for motion planning applications.

\section{Conclusion}
\label{sec:conclusion}

We introduce StyledStreets, a framework for photorealistic urban scene editing that achieves instruction-driven style transfer while preserving spatial-temporal consistency. By integrating hybrid style-geometry embeddings, uncertainty-aware diffusion guidance, and unified parametric modeling, our method overcomes key limitations in neural scene editing. Experimental validation on the Waymo Open Dataset \cite{waymo2020} demonstrates state-of-the-art performance, with quantitative improvements of +2.15 dB PSNR in vehicle reconstruction and 18\% Chamfer Distance reduction compared to baseline approaches. Qualitative results (Figs.~\ref{fig:streetgaussian_ours}, \ref{fig:depth_unified_model}) confirm faithful style transfer across extreme environmental conditions while maintaining geometric fidelity critical for autonomous systems. The joint pose optimization and multi-view training strategies enable reliable simulation for vehicle-mounted camera arrays, establishing new capabilities for urban digital twins. Future work will explore real-time editing optimizations and physics-aware style transfer for enhanced practical deployment.

\newpage
{
    \small
    \bibliographystyle{ieeenat_fullname}
    \bibliography{main}
}

\clearpage
\setcounter{page}{1}
\maketitlesupplementary

\section{Multi-View Training}
\label{sec:mv-training}

Multi-view training provides three synergistic benefits: variance reduction via anti-correlated perspective gradients; signal amplification through directional consensus on scene geometry; and improved local minima escape from gradient magnitude scaling. These jointly explain the empirical improvements quantified in Table~\ref{tab:results}.

\paragraph{Preliminary.} Our multi-view training strategy prioritizes temporally adjacent frames (frame gap $\leq$3) over distant observations (frame gap $>$10) through exponential sampling weights $w(t) = \exp(-\lambda|t|)$. Each minibatch contains $M$ views from different cameras $\{\theta_m\}_{m=1}^M$ across adjacent timestamps $t \pm \Delta t$, ensuring spatial-temporal coherence while maintaining pose diversity. This design enables gradient aggregation across synchronized perspectives without temporal aliasing.

3D Gaussian Splatting represents scenes through explicit anisotropic 3D Gaussians $\mathcal{G} = \{g_i\}_{i=1}^N$, where each Gaussian $g_i = (\mu_i, \Sigma_i, c_i, o_i)$ comprises four components: positional mean $\mu_i \in \mathbb{R}^3$ in world coordinates; covariance $\Sigma_i \in \mathbb{S}_{++}^3$ controlling spatial extent; view-dependent color $c_i \in \mathbb{R}^3$ via spherical harmonics basis $\psi(\omega): \mathbb{S}^2 \to \mathbb{R}^3$ with view direction $\omega$; and opacity $o_i \in [0,1]$ governing light transmission. 

The differential rendering equation accumulates contributions along ray $r(t) = o + t\mathbf{d}$ through these Gaussians:
\begin{align}
    \mathbf{C}(r) &= \sum_{i=1}^N c_i \alpha_i \prod_{j=1}^{i-1}(1-\alpha_j), \\
    \alpha_i &= o_i \exp\left(-\frac{1}{2}\Delta x_i^T\Sigma_i^{-1}\Delta x_i\right),
\end{align}
where $\Delta x_i = x_i-\mu_i$ denotes positional offset from Gaussian mean to sample point $x_i$. This explicit representation enables real-time $\mathcal{O}(N)$ rendering via tile-based rasterization.

\subsection{Multi-View Gradient Aggregation}

For Gaussian primitive parameters $\theta \in \{\mu_i, \Sigma_i, o_i, c_i\}$, the full gradient chain through the rendering equation aggregates multi-view photometric gradients:
\begin{equation}
    \frac{\partial\mathcal{L}_{\text{MV}}}{\partial\theta} = \frac{1}{M}\sum_{m=1}^M\sum_{p\in\mathcal{P}_m} \underbrace{\frac{\partial\mathcal{L}_m}{\partial\mathbf{C}_m(p)}}_{\text{Image gradient}} \cdot \underbrace{\frac{\partial\mathbf{C}_m(p)}{\partial\alpha_i}}_{\text{Blending}} \cdot \underbrace{\frac{\partial\alpha_i}{\partial\theta}}_{\text{Primitive physics}},
\end{equation}
where $p$ indexes pixels in view $m$'s pixel set $\mathcal{P}_m$. The image gradient term $\partial\mathcal{L}_m/\partial\mathbf{C}_m(p)$ measures photometric error sensitivity, while the blending gradient $\partial\mathbf{C}_m(p)/\partial\alpha_i$ captures alpha compositing effects. The primitive physics gradient $\partial\alpha_i/\partial\theta$ encodes each Gaussian's geometric response.

\subsubsection{Position Gradients} For positional mean $\mu_i$, gradients exhibit pairwise occlusion effects:
\begin{align}
    \frac{\partial\mathbf{C}_m}{\partial\mu_i} &= \sum_{j=i}^N c_j\frac{\partial\alpha_j}{\partial\mu_i}\prod_{k=1}^{j-1}(1-\alpha_k) \nonumber \\
    &- \sum_{j=i+1}^N c_j\alpha_j\frac{\partial\alpha_i}{\partial\mu_i}\prod_{k=1,k\neq i}^{j-1}(1-\alpha_k), \\
    \frac{\partial\alpha_i}{\partial\mu_i} &= \alpha_i\Sigma_i^{-1}(x_i - \mu_i).
\end{align}
The second summation arises from $\mu_i$'s impact on subsequent Gaussian transmittance.

\subsubsection{Covariance Gradients} Covariance $\Sigma_i$ gradients follow similar occlusion dynamics:
\begin{align}
    \frac{\partial\mathbf{C}_m}{\partial\Sigma_i} &= \sum_{j=i}^N c_j\frac{\partial\alpha_j}{\partial\Sigma_i}\prod_{k=1}^{j-1}(1-\alpha_k) \nonumber \\
    &- \sum_{j=i+1}^N c_j\alpha_j\frac{\partial\alpha_i}{\partial\Sigma_i}\prod_{k=1,k\neq i}^{j-1}(1-\alpha_k), \\
    \frac{\partial\alpha_i}{\partial\Sigma_i} &= \frac{1}{2}\alpha_i\left[\Sigma_i^{-1}(x_i-\mu_i)(x_i-\mu_i)^T\Sigma_i^{-1} - \Sigma_i^{-1}\right].
\end{align}

\subsubsection{Opacity Gradients} Opacity $o_i$ gradients balance local density and global visibility:
\begin{align}
    \frac{\partial\mathbf{C}_m}{\partial o_i} &= c_i\frac{\partial\alpha_i}{\partial o_i}\prod_{k=1}^{i-1}(1-\alpha_k) \nonumber \\
    &- \sum_{j=i+1}^N c_j\alpha_j\frac{\partial\alpha_i}{\partial o_i}\prod_{k=1,k\neq i}^{j-1}(1-\alpha_k), \\
    \frac{\partial\alpha_i}{\partial o_i} &= \exp\left(-\frac{1}{2}(x_i-\mu_i)^T\Sigma_i^{-1}(x_i-\mu_i)\right).
\end{align}

\subsubsection{Color Gradients} Spherical harmonics coefficients $c_i$ exhibit view-local gradients:
\begin{equation}
    \frac{\partial\mathbf{C}_m}{\partial c_i} = \alpha_i\prod_{k=1}^{i-1}(1-\alpha_k)\psi(\omega_m),
\end{equation}
where $\psi(\omega_m)$ encodes view direction $\omega_m$ for camera $m$.

\subsection{Variance Bound Analysis}
The multi-view gradient variance $\text{Var}\left[\frac{\partial\mathcal{L}_{\text{MV}}}{\partial\theta}\right]$ follows:
\begin{align}
    \text{Var} \leq \frac{1}{M^2}\left(M\sigma^2 + 2\sum_{1\leq i<j\leq M}\text{Cov}\left(\frac{\partial\mathcal{L}_i}{\partial\theta},\frac{\partial\mathcal{L}_j}{\partial\theta}\right)\right),
\end{align}
with covariance terms becoming negative under perspective-aware filtering. Occlusion constraints and z-buffering induce anti-correlated gradients between views, suppressing variance.

\subsection{Directional Consistency}
Gradient direction alignment emerges from:
\begin{equation}
    \left\|\frac{\partial\mathcal{L}_{\text{MV}}}{\partial\theta}\right\|_2^2 = \frac{1}{M^2}\left(\sum_{m=1}^M\left\|\frac{\partial\mathcal{L}_m}{\partial\theta}\right\|_2^2 + 2\sum_{i<j}\left\langle\frac{\partial\mathcal{L}_i}{\partial\theta},\frac{\partial\mathcal{L}_j}{\partial\theta}\right\rangle\right),
\end{equation}
where coherent structural priors amplify directional consensus ($\langle\cdot,\cdot\rangle > 0$) while view-dependent effects average out. This yields $\mathcal{O}(\sqrt{M})$ magnitude growth for aligned gradients, accelerating convergence while maintaining stability.

\section{Depth Rendering in World Coordinates}
\label{sec:depth_rendering}

The rendered depth map $\mathbf{D}_m \in \mathbb{R}^{H \times W}$ for camera view $m$ combines projective geometry and uncertainty-aware Gaussian contributions. For pixel $p$ with ray direction $\mathbf{d}_p$ originating at camera center $\mathbf{o}_m$, the depth computation involves:

\begin{align}
z_i^{cam} &= [\mathbf{R}_{ego}^{m,t} \mu_i^{world} + \mathbf{t}_{ego}^{m,t}]_z, \\
\mathbf{D}_m(p) &= \sum_{i=1}^N z_i^{cam} \alpha_i \prod_{j=1}^{i-1}(1-\alpha_j),
\end{align}

where $\alpha_i$ follows the same compositing weights as color rendering (see \secref{sec:mv-training}). The camera-space covariance $\Sigma_i^{cam} = \mathbf{R}_{ego}^{m,t} \Sigma_i^{world} (\mathbf{R}_{ego}^{m,t})^\top$ ensures perspective-correct scaling of Gaussian footprints during projection.

World coordinates are recovered through inverse transformation:

\begin{equation}
\mathbf{D}_m^{world}(p) = (\mathbf{R}_{ego}^{m,t})^{-1} \left( \mathbf{D}_m(p) \cdot \mathbf{d}_p + \mathbf{o}_m - \mathbf{t}_{ego}^{m,t} \right).
\end{equation}

This preserves metric scale using known camera extrinsics $\mathbf{T}_{ego}^{m,t} = (\mathbf{R}_{ego}^{m,t}, \mathbf{t}_{ego}^{m,t})$. The z-buffering mechanism in \eqref{eq:rendering_equation} maintains consistent depth ordering across views through shared world coordinates, enabling cross-camera geometric consistency under ego-motion.

\section{More Qualitatives}
\label{sec:more_qualitatives}

\begin{figure*}[!t]
\centering
\includegraphics[width=\linewidth]
{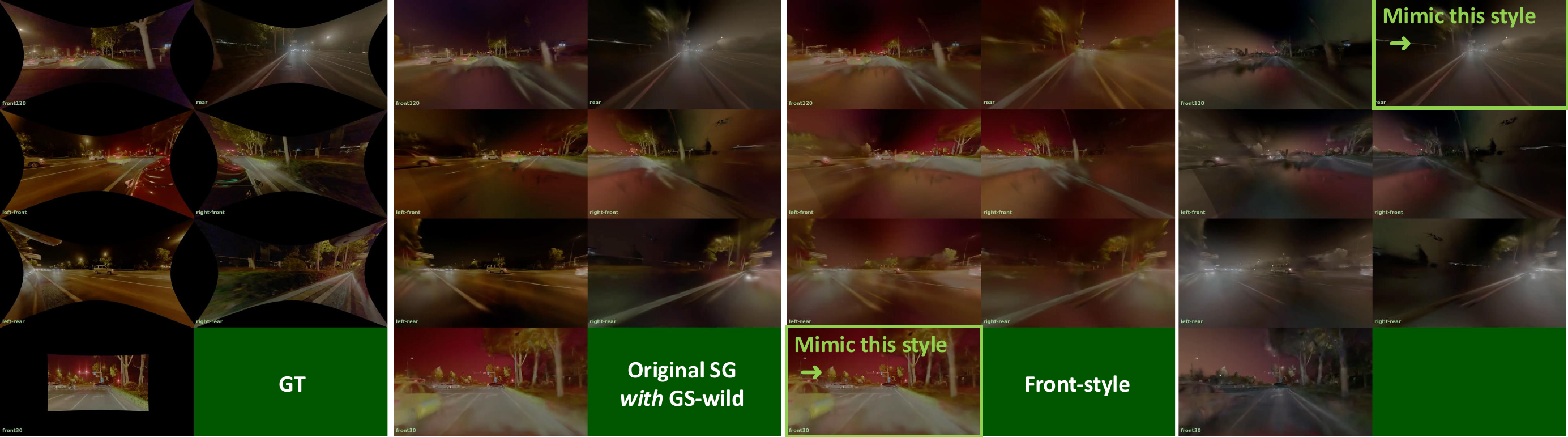}
\caption{Assume that our street GS model is trained with 7 cameras with different setups as shown in (a), \textit{e.g.} exposure parameters, FoV, the style of all 7 cameras could be parameterized with $\mathbf{e}_i^{\text{cam}}$. So each unique camera setup can be adopted for all cameras as shown in (c), (d) when we change the $\mathbf{e}_i^{\text{cam}}$ accordingly.}
\label{fig:cameras_appearance_embedding}
\end{figure*}

\begin{figure*}[!t]
\centering
\includegraphics[width=\linewidth]
{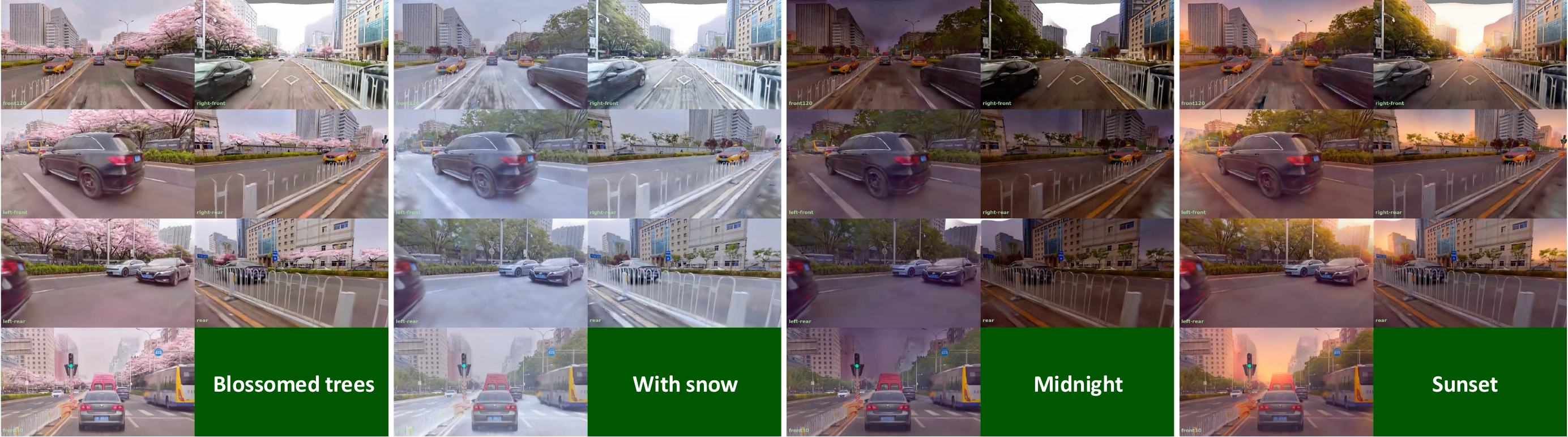}
\caption{Change the season and day time with the help of $\mathbf{e}^{\text{style}}(T)$.}
\label{fig:seasons-daytime}
\end{figure*}

\begin{figure*}[!t]
\centering
\includegraphics[width=\linewidth]
{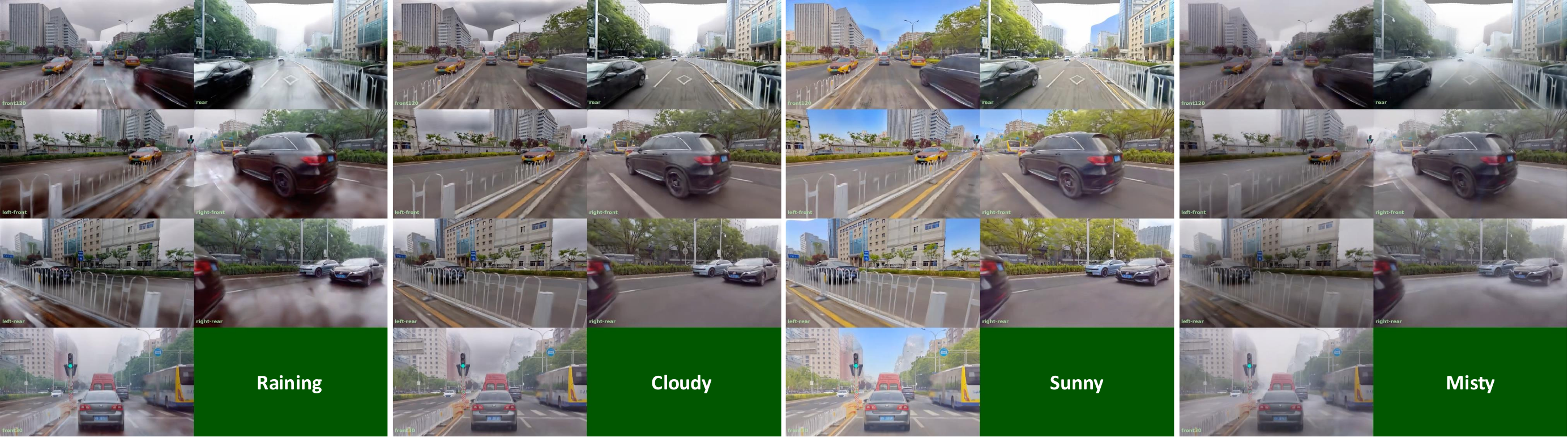}
\caption{Adapting some special weathers such as raining, cloudy, sunny, and misty with the help of different $\mathbf{e}^{\text{style}}(T)$.}
\label{fig:weather}
\end{figure*}

The original ground truth data in \figref{fig:cameras_appearance_embedding} is collected by 7 cameras mounted on the egocentric vehicle. \figref{fig:cameras_appearance_embedding} demonstrates appearance adjustments with different $\mathbf{e}_i^{\text{cam}}$ for different camera setups for in-the-wild data. \figref{fig:seasons-daytime} and \figref{fig:weather} shows the appearance change regarding to different $\mathbf{e}_i^{\text{style}}(T)$.

\begin{figure*}[!t]
\centering
\includegraphics[width=\linewidth]{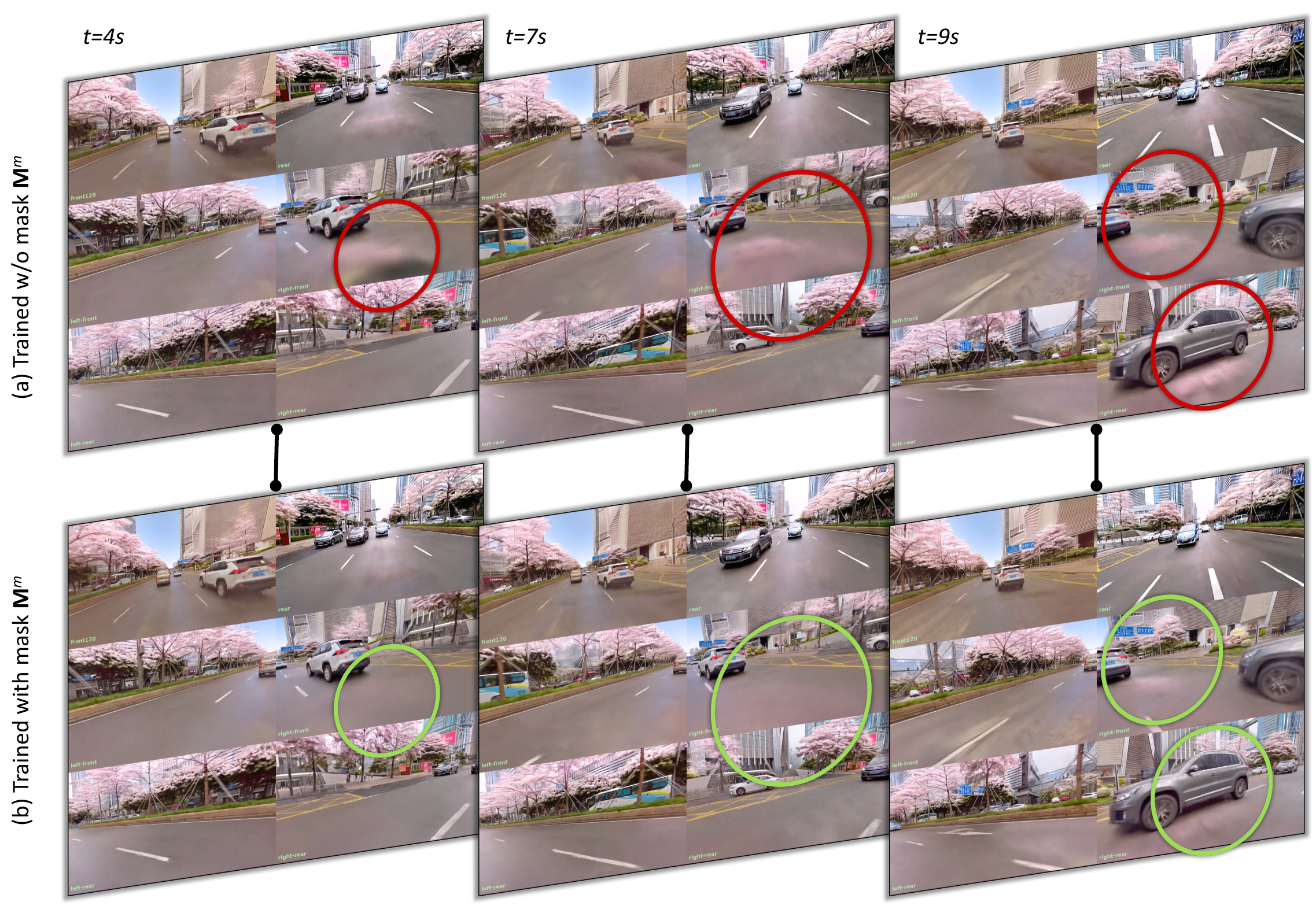}
\caption{Qualitative results of 3 randomly selected frames trained with our proposed method trained with 6 cameras, investigating the advantages introduced by DINO masks $\mathbf{M}$ to mitigate ambiguous 2D diffusion supervision. The wrongly assigned flower patterns on the ground in subfigure (a) are corrected as shown in (b).}
\label{fig:dino_liauto}
\end{figure*} 

\begin{figure*}[!t]
\centering
\includegraphics[width=\linewidth]
{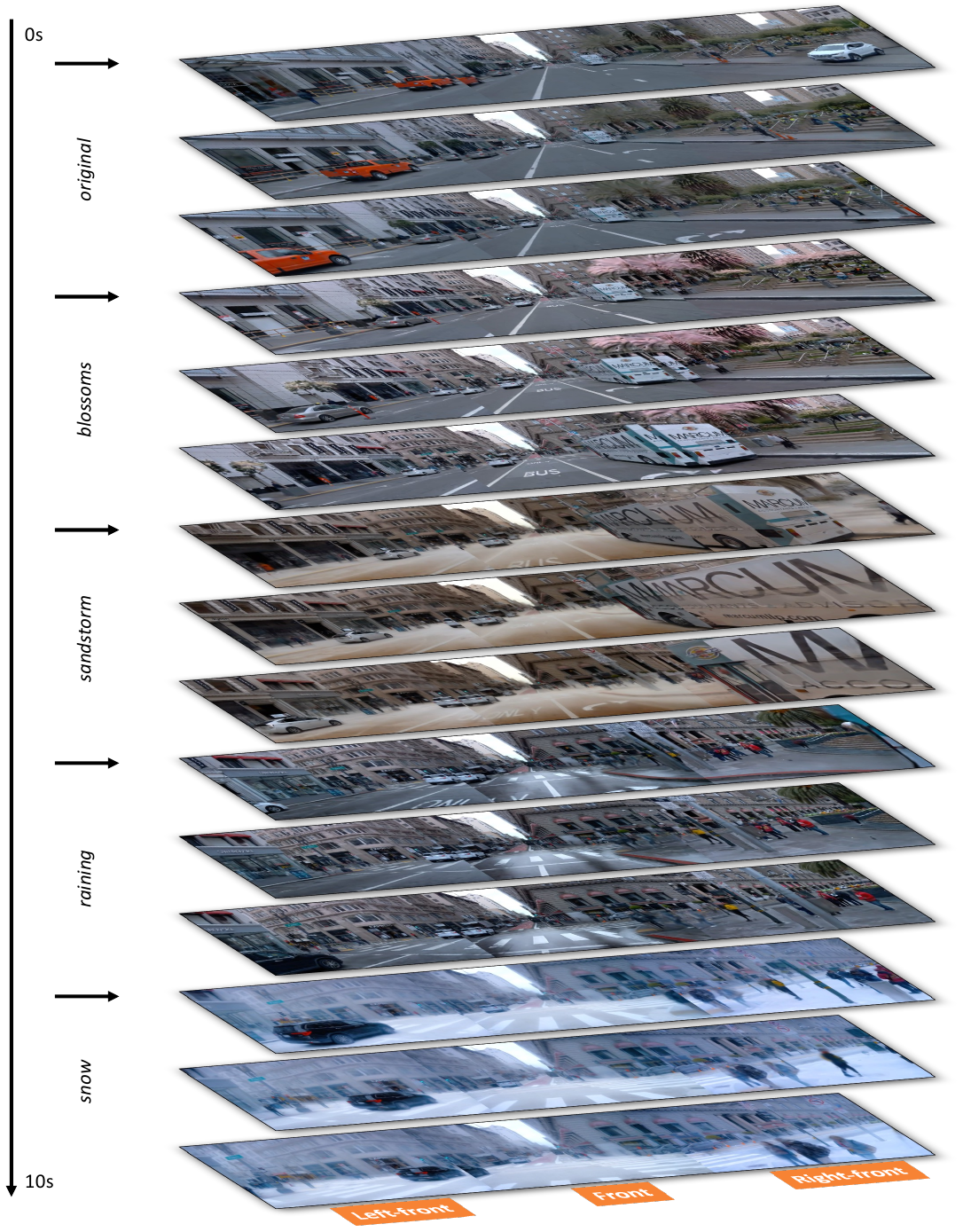}
\caption{Some selected frames between 0s and 10s trained in the Waymo dataset for 3 cameras, simulated with the unified model presenting 5 styles.}
\label{fig:video_5styles}
\end{figure*}

\section{3D Consistency in 2D Diffusion Models}
\label{sec:3d_consistency}
A well-trained image or video diffusion model $\mathcal{D}$ achieves \textit{3D consistency} when generating novel views or styles from 2D inputs if, for any pair of camera poses $\{\mathbf{P}_k, \mathbf{P}_l\}$ or stylization parameters $\{T_k, T_l\}$, it satisfies the dense correspondence
\begin{equation}
    \mathcal{D}\left(\mathbf{I}_k^{\text{gt}}; \mathbf{P}_k \to \mathbf{P}_l \text{ or } T_k \to T_l\right) = \mathbf{I}_l^{\text{gt}},
\end{equation}

where $\mathbf{I}_l^{\text{gt}}$ denotes the expected image under pose $\mathbf{P}_l$ or style $T_l$. This fundamental property implies complementary geometric constraints through sparse 3D correspondence as well.

\paragraph{Sparse 3D Correspondence}
For any ground truth image pair $(\mathbf{I}_k, \mathbf{I}_l)$, their 2D keypoints $\{\mathbf{x}_k^i\}$ and $\{\mathbf{x}_l^j\}$ must satisfy
\begin{equation}
    \forall i \in \mathcal{S},\ \exists j \in \mathcal{S}:\ \pi^{-1}(\mathbf{x}_k^i; \mathbf{P}_k, d_k^i) = \pi^{-1}(\mathbf{x}_l^j; \mathbf{P}_l, d_l^j),
    \label{equ:3d_correspondence_mvs}
\end{equation}

where $\pi^{-1}$ denotes back-projection using depth $d$ and camera parameters, and $\mathcal{S}$ indexes shared structural elements. This geometric consistency forms the foundational assumption for convergent 3D reconstruction in Multi-View Stereo (MVS) paradigms like NeRF, NeuS, and 3D Gaussian Splatting.

The generated outputs $\mathcal{D}(\mathbf{I}_k)$ and $\mathcal{D}(\mathbf{I}_l)$ must preserve this 3D correspondence through
\begin{equation}
    \forall i \in \mathcal{S},\ \exists j \in \mathcal{S}:\ \pi^{-1}(\mathbf{x}_k^i; \mathbf{P}_k, d_k^i) = \pi^{-1}(\mathbf{x}_l^j; \mathbf{P}_l, d_l^j),
\end{equation}

where $\{\mathbf{x}_k^i, \mathbf{x}_l^j\}$ represent equivalent local structures in the synthesized views. The depth estimates $d$ and camera parameters must remain consistent across both observation and generation domains. Our result in ~\figref{fig:novel_view} demonstrates that our model preserves such 3D consistency.

\end{document}